\documentclass[electronic]{vgtc}             

\ifpdf
  \pdfoutput=1\relax                   
  \usepackage{graphicx}                
  \DeclareGraphicsExtensions{.pdf,.png,.jpg,.jpeg} 
\else
  \usepackage{graphicx}                
  \DeclareGraphicsExtensions{.eps}     
\fi%

\usepackage{microtype}                 
\PassOptionsToPackage{warn}{textcomp}  
\usepackage[full]{textcomp}                  
\usepackage{mathptmx}                  
\usepackage{times}                     
\usepackage{cite}                      
\usepackage{tabu}                      
\usepackage{soul}
\usepackage{url}
\usepackage[hidelinks]{hyperref}
\usepackage[utf8]{inputenc}
\usepackage{graphicx}
\usepackage{subfig}
\usepackage{amsmath}
\usepackage{newtxmath} 
\usepackage{newtxtext}
\usepackage{bm}
\usepackage{booktabs}
\usepackage{enumitem}
\usepackage{xcolor}

\onlineid{1050}

\vgtccategory{Research}

\vgtcinsertpkg

\newcommand{\myvspace}{\vspace{-6pt}}
\newcommand{\noindentbf}[1]{\vspace{3pt}\noindent\textbf{#1}}

\title{Contrastive Identification of Covariate Shift in Image Data}

\author{
Matthew L. Olson,
Thuy-Vy Nguyen,
Gaurav Dixit,
Neale Ratzlaff,
Weng-Keen Wong, and
Minsuk Kahng\thanks{e-mail:
\{olsomatt, nguythu2, dixitg, ratzlafn, wongwe, minsuk.kahng\}\allowbreak@oregonstate.edu
}}
\affiliation{\scriptsize Oregon State University}

\abstract{
Identifying covariate shift is crucial for making machine learning systems robust in the real world and for detecting training data biases that are not reflected in test data. However, detecting covariate shift is challenging, especially when the data consists of high-dimensional images, and when multiple types of localized covariate shift affect different subspaces of the data.
Although automated techniques can be used to detect the existence of covariate shift, our goal is to help \textit{human} users characterize the extent of covariate shift in large image datasets with interfaces that seamlessly integrate information obtained from the detection algorithms. 
In this paper, we design and evaluate a new 
visual interface
that facilitates the comparison of the local distributions of training and test data. 
We conduct a quantitative user study on multi-attribute facial data to compare two different learned low-dimensional latent representations (pretrained ImageNet CNN vs. density ratio) and two user analytic workflows (nearest-neighbor vs. cluster-to-cluster).
Our results indicate that the latent representation of our density ratio model, combined with a nearest-neighbor comparison, is the most effective at helping humans identify covariate shift.
}

\begin{document}

\maketitle

\section{Introduction}

One of the common problems that plague deployed machine learning (ML) systems is covariate shift \cite{Shimodaira00}, which occurs when the input feature distribution $P(\bf{X}$) changes between training and testing phases, but the conditional distribution of the response given the features $P(Y|\bm{X})$, remains the same. For example, an image recognition system trained during sunny days may not be effective on cloudy days. By not accounting for covariate shift, ML systems can lack robustness when they encounter "unknown unknowns" \cite{lakkaraju2017identifying} during deployment and are therefore vulnerable to bias in the training data. 

Although automated algorithms can be effective at detecting covariate shift (e.g., Chapters 6-10 in \cite{DatasetShift09}), 
humans still need to be involved in the process for several reasons: 
first, it is an important task for people to detect if the data distribution has changed enough to affect a ML system that has been deployed. 
If the data exhibits bias or shift, they need to know it, so that they can take further actions.
Second, it is possible that multiple types of localized covariate shift are occurring in the dataset, with each type affecting a different subspace of the overall feature space. These localized covariate shifts can be challenging for an algorithm to identify and past work has shown that humans can sometimes be better than machines at detecting these problem areas \cite{Attenberg15}. %
Third, a human is often ultimately needed to identify the cause of the shift and to fix the problem.

Identifying covariate shift from image data, among the many types of data used in ML, is more challenging because the data is high-dimensional and the original (pixel) feature space is less human-interpretable. 
Can visualization help human users to identify and characterize how test set images are different from training set images (e.g., face images in the training set have no glasses while the test set has some)~\cite{spinner2019explainer}?
One possible approach may be to visualize training and test distributions side-by-side (i.e., juxtaposition) using dimensionality reduction methods (e.g., t-SNE)~\cite{arendt2020parallel} 
and show each data point as an image thumbnail~\cite{wexler2019if,chen2020oodanalyzer}. 
However, the scale of modern image datasets makes it difficult because
we cannot easily show many images on the projected space~\cite{chen2020oodanalyzer,hohman2018visual}. 
Instead of visualizing the distributions of the entire training and test datasets \emph{globally}, we aim to intelligently show only \emph{local} regions of the space, where the locality is informed by the detection algorithm.
For example, given a test set image highly ranked by a shift detection algorithm (i.e., deviated from training set distribution), a visualization may show that many of its similar test images (i.e., local neighborhood) share a characteristic (e.g., many faces with sunglasses) while the similar training images do not (e.g., no faces with sunglasses).

In this paper, we design and evaluate a new visual analysis interface for human users to identify covariate shift in image data.
Although there exist several visualization work for detecting some types of dataset shift~\cite{chen2020oodanalyzer,yang2020diagnosing,Schneider20,wang2020conceptexplorer}, we advance beyond the existing work in two aspects.
First, our interface is designed to facilitate \textit{contrastive} analysis between two different distributions for local regions~\cite{gleicher2017considerations,arendt2020parallel}, which is a key to covariate shift detection task. We design a novel \textit{side-by-side histogram} view for comparing two sets of images in a selected local region and characterizing shifts.
Second, while past work often uses the raw image features in embedding into two-dimensional (2D) space, we integrate the internal latent representation information of detection algorithms into computing similarities between images, which more accurately presents distribution differences. 

We address the following two key research questions which we investigate in our 2$\times$2 quantitative user study: 
\vspace{-2pt}
\paragraph{(RQ1) Which learned lower-dimensional latent representation is the most useful for humans to detect covariate shift?} Comparing two high-dimensional representations requires some form of dimensionality reduction. We thus compare two lower-dimensional latent representations learned by deep neural networks. The first representation is a commonly used but effective baseline obtained from a pre-trained ImageNet Convolutional Neural Net (CNN) \cite{huh2016makes}. For the second latent representation, we performed an empirical evaluation and found that the most effective latent representation for a ML algorithm to detect covariate shift is learned through a density ratio estimation (DRE) neural network \cite{Nam15}. 
We want to evaluate how effective it can be for humans.
\vspace{-2pt}
\paragraph{(RQ2) Which analytic workflow is the most effective at identifying covariate shift?} 
The side-by-side visualization is designed to work for analyzing the local regions of the feature space. We explore two different user workflows for selecting local regions in discovering covariate shift: (1) a nearest-neighbor approach: a user picks an image that a detection algorithm estimates to be a likely outlier and the user sees similar images; and (2) a cluster-to-cluster approach: a user is presented with a set of clusters and examines each cluster. 

\section{Related Work}
Dataset shift \cite{DatasetShift09} is a broad topic covering many ways test data can be different from training data. 
Schneider et al.~\cite{Schneider20} presented a visualization design space for dataset shift and a tool for comparing multi-dimensional feature distributions. 
In terms of specific types of dataset shift, \textit{concept drift} is a topic that has garnered some attention \cite{yang2020diagnosing,wang2020conceptexplorer}. Concept drift occurs when the relationship between the response variable and the features (i.e., $P(Y|\bm{X}))$ changes between training and testing. However, covariate shift addressed in our paper is fundamentally different from concept drift as it detects differences in the training and test feature distributions. Another category of approaches deal with detecting ``\textit{unknown unknowns}'' (UUs) \cite{lakkaraju2017identifying,liu2020towards}, which are data instances incorrectly classified with high confidence because the training data is missing entire subclasses, thereby causing blind spots \cite{Attenberg15}. Detecting UUs focuses on finding misclassified instances with high confidence. In contrast, our focus on covariate shift ignores classifier confidence and only looks at differences between training and test data distributions for image data.
The most closely related work is OoDAnalyzer \cite{chen2020oodanalyzer} for detecting out-of-distribution (OoD) images. It first detects OoD instances using a deep ensemble, then the data instances from the original feature space are mapped to a 2D space through a grid-based layout algorithm.
Our approach instead aims to enable users to characterize localized covariate shifts affecting a subspace of the features and explores 
to use different lower dimensional latent spaces extracted from shift detection models rather than the original feature space.

\section{Latent Representations of Shift Detection Models}

\begin{figure}[t]%
    \centering
    \includegraphics[width=.95\linewidth]{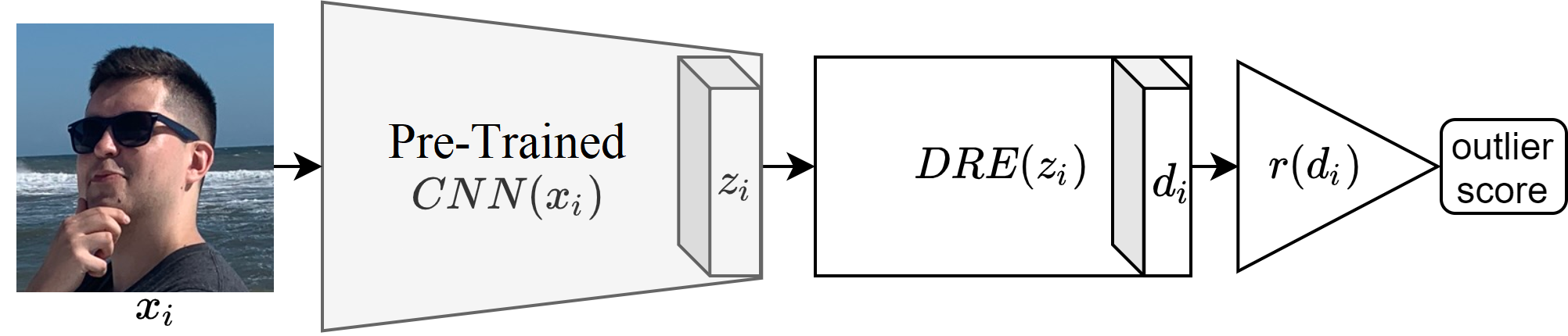}
    \vspace{-5pt}
    \caption{Our ML architecture for computing an outlier score for an image using a pre-trained CNN and our actively trained DRE-based model. We highlight the two vector latent representations of interest, $\bm{z_i}$ and $\bm{d_i}$, which are compared in our user study.
    }
    \label{fig:dre_model}%
    \myvspace
\end{figure}

\label{sec:latentrepresentations}
We investigated a variety of latent representations for our task (see supplemental material for more details) and found that the latent representation learned by a \textit{density ratio estimation} (DRE) algorithm performed the best. In this section, we introduce this DRE algorithm, which assigns an \textit{outlier} score to each test instance (lower value means it is unlikely to be drawn from the training set distribution).

Let $\bm{x}_i$ be the raw input features (i.e., the pixels of the image) of the $i$-th instance in a dataset.
Let $\bm{z}_i = \text{CNN}(\bm{x}_i)$ be the learned latent representation from a pre-trained Convolutional Neural Network; for instance, this latent representation is the penultimate layer of the pretrained InceptionNet CNN model \cite{szegedy2015going}. 
A superscript of $\text{tr}$ or $\text{te}$ denotes the training or test dataset respectively. For instance, $x_i^{\text{tr}}$ indicates the $i$-th training data instance's features.

Next, in \autoref{fig:dre_model}, $\bm{d}_i = \text{DRE}(\bm{z}_i)$ is the representation learned when training a DRE-based model, where the density ratio $r(\bm{d}_i) = P^{\text{tr}}(\bm{d}_i) / P^{\text{te}}(\bm{d}_i)$ is the ratio of the training density divided by the test density. We use $r(\bm{d}_i)$ for determining the outlier score, where the lower the value, more likely the instance will be an outlier. We use the Kullbeck-Liebler Importance Estimation Procedure (KLIEP) as the DRE method because it outperformed other DRE methods in our preliminary investigations. 
The KLIEP loss is defined as follows:
\vspace{-6pt}
\begin{equation}
L_{\text{KLIEP}} = \frac{1}{n_{\mathrm{te}}} \sum_{j=1}^{n_{\mathrm{te}}} r\left(\boldsymbol{d}_{j}^{\mathrm{te}}\right)-\frac{1}{n_{\mathrm{tr}}} \sum_{i=1}^{n_{\mathrm{tr}}} \ln \left(r\left(\boldsymbol{d}_{i}^{\mathrm{tr}}\right)\right),
\vspace{-3pt}
\end{equation}
where $r(\bm{d}) = \log(\exp(W \bm{d} + b)+1)$ to ensure non-negativity.

\section{Visual Interface and Analytic Workflow Design}

\begin{figure}[!t]
        \includegraphics[trim=0.1in 0 1.2in 0,clip,width=\linewidth]{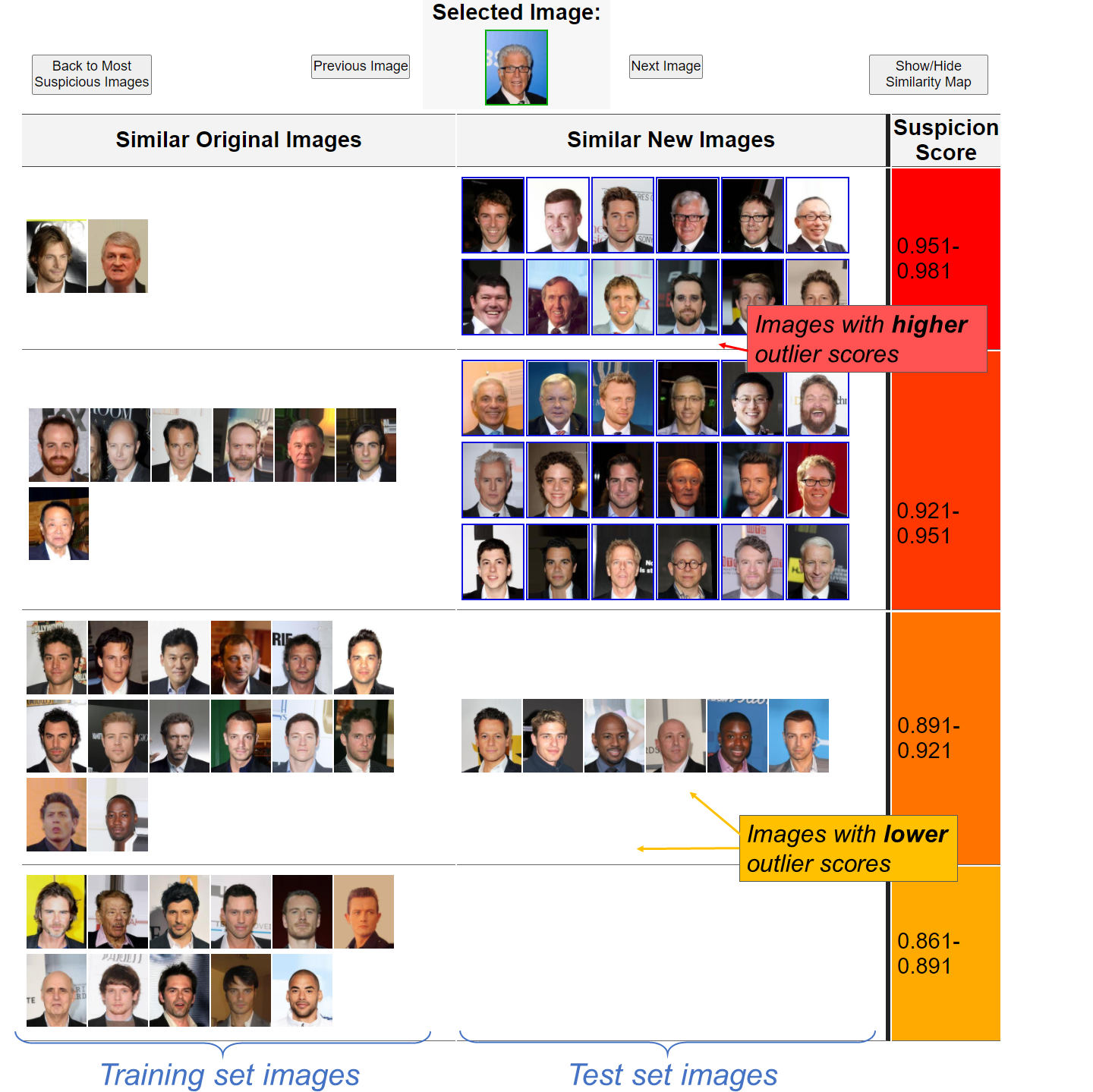}
        \vspace{-14pt}
        \caption{Our \textit{side-by-side histogram} is designed to help users easily compare the distributions of training and test set images (here we used the term ``original'', ``new'', and ``suspicion score''-- instead of ``training set'', ``test set'', and ``inverse density ratio''-- for non-expert users).
        It shows both training and test set images (on the left and right side, respectively) that are close to a selected image (shown at top).
        In this example, 
        the similar test set images on the right side have more images with high outlier scores compared to the training set on the left, and
        only the right side includes face images with smiles and neckties while the left side does not include such images.
        }\myvspace
        \label{fig:nn_table}
\end{figure}

This section describes two versions of workflows and associated visual interfaces for covariate shift identification.

\noindentbf{Typical ML-only approach.}
A typical way of detecting outliers or shifts from image data (without visualization) is by having a large list of test images sorted by outlier score, then walking through each image in the list one by one.
This method falls short in enabling users to compare a test set against a training set and finding patterns in test data for the way covariate shift may occur.

\noindentbf{Typical VIS-only approach.}
On the other hand, one approach to visually comparing two distributions is using two 2D projected views side-by-side. However, it does not scale especially if we want to show individual images directly on the projected view.

We combine the ML and VIS approaches by allowing users to select local regions of the data space with the help of shift detection algorithms and visually compare the training and test set distributions for these regions. In selecting local regions, we consider two workflows:
(1) \textit{nearest-neighbor} and (2) \textit{cluster-to-cluster}.

\subsection{Nearest Neighbor User Workflow}

In the first user workflow,
a user begins with a list of images sorted by shift scores and examines images one by one, just like the typical ML-only without-visualization approach of analyzing results from outlier detection algorithms. 
The difference is that the user examines each image with
a new visual contrastive interface that shows the neighborhood of the selected image both for training and test sets.

\noindentbf{New contrastive visualization for shift identification.}
Once a user selects an image,
they are provided with our new \textit{side-by-side histogram} visualization (shown in \autoref{fig:nn_table}).
For a selected image, shown at the top,
the visualization displays two vertical histograms: one for training set (shown on the left) and the other for test set (on the right). 
Its vertical bins are computed using a shift detection model's prediction of covariate shift, normalized and sorted over all data. We found sorting to be important for drawing a user's attention to images most likely to contain a shift.
This histogram of images not only shows an outlier score distribution of images, but also displays individual examples of images,
motivated by the \textit{unit visualization} technique~\cite{park2017atom} which represents individual data points within the context of aggregate statistics.
For example, in \autoref{fig:nn_table}, users can see that test data (on the right side) has more high scored outliers (more images on the top rows) compared to training data (on the left), and
faces with glasses, smiles, and neckties appear only on the right side---while none of these attributes occur on the left.

\noindentbf{Detailed setup used in user study.} 
For our study, we set the number of histogram bins to be 5 and programatically search for a distance where at least 100 images are chosen for half of the selected images, and set a minimum threshold of number of images to ten for the other half. 
Distance is calculated by the Frobenius norm in the latent space between the selected image and all others, and neighbors are instances with a small distance.
We additionally provide participants with a 2D projection of the latent representation of the test data using  UMAP~\cite{becht2019dimensionality} with default parameters,
to show a global picture of the feature space for all images.
The interface also supports interactions.
For example, when users interacting with the histogram view, the 2D projection view is updated to highlight the selected image and the associated test images to help them see their location in a global picture. A participant can navigate to a new histogram by clicking an image in the histogram view or a data point on the projected view.

\begin{figure}[!tb]
\centering
        \includegraphics[width=.9\linewidth]{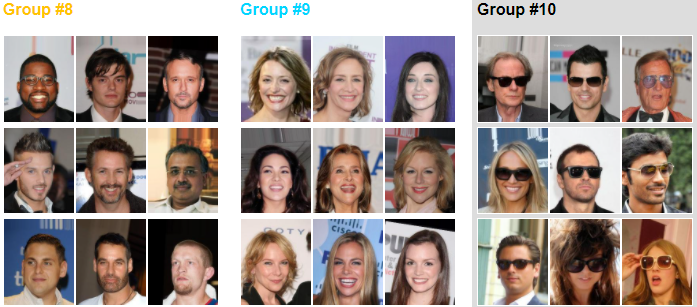}
        \vspace{-7pt}
        \caption{Front page of the interface for the cluster-to-cluster exploration.
        Images in Cluster \#10 include face images with sunglasses. We replaces the term ``clusters'' with ``groups'' for non-expert users.
        }
        \myvspace
        \vspace{9pt}
        \label{fig:cluster_groupings}
\end{figure}

\begin{figure}[!tb]
\centering
        \includegraphics[width=\linewidth]{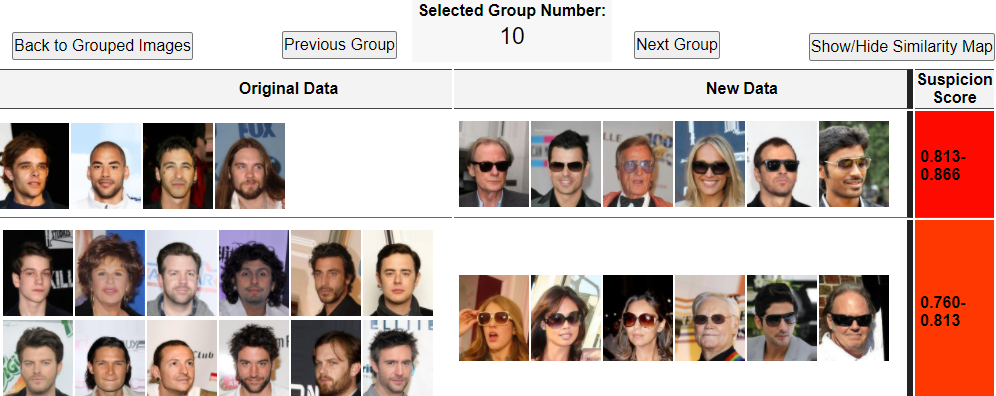}
        \vspace{-16pt}
        \caption{The \textit{side-by-side histogram} with the cluster workflow, showing only the top portion of the interface.
        }\myvspace
        \label{fig:cluster_histogram_table}
\end{figure}

\subsection{Cluster-to-Cluster User Workflow}
We design the other workflow, \textit{cluster-to-cluster}, to help users analyze a large number of images without having to  select each image one by one.
Instead of presenting a sorted list at the beginning, this version of the interface presents a set of clusters with representative images, as shown in \autoref{fig:cluster_groupings}. Users can choose one of the clusters to see the corresponding side-by-side histogram for the selected cluster (an example shown in \autoref{fig:cluster_histogram_table}).
The visualization looks very similar to that for the nearest neighbor workflow. The main difference being the nearest neighbor shows a selected image at the top while the cluster-to-cluster view simply shows a cluster ID.

\noindentbf{Detailed setup used in user study.}
A participant is first shown with 10 clusters that potentially have more outliers, each with its top nine outlier images as its representatives (as in \autoref{fig:cluster_groupings}). 
To determine the 10 clusters, we first create 100 clusters from the test data by using an agglomerative clustering algorithm~\cite{mullner2011modern} over the latent representation of test set. We then compute the average outlier score for each cluster and select the 10 clusters with highest outlier score.

For the side-by-side histogram for a cluster, a maximum of 50 images in the cluster are shown for the test set, and an equal number of training set images are selected for effective comparison of two distributions, where the training images are the closest images to the cluster's centroid in the latent representation space. 

\section{User Study}
We conducted an online human subject study with a 2$\times$2 design,
to answer the two research questions we asked earlier in Section 1: 

\begin{enumerate}[leftmargin=0.4in,topsep=1pt,parsep=0pt,partopsep=0pt,itemsep=2pt,label=\textbf{RQ\arabic*.}]
    \item Which learned lower-dimensional latent representation is more effective for humans to detect covariate shift? (i.e., pre-trained ImageNet CNN vs. density ratio) and 
    \item Which analytic workflow is more effective at identifying covariate shift? (i.e., nearest-neighbor vs. cluster-to-cluster).
\end{enumerate}

\subsection{Study Design}

\noindentbf{Participants.}
We recruited 60 unique participants using university mailing lists.
The average age was 24 with a standard deviation of 6.
There were 42 male, 17 female, and 1 gender non-conforming participants.
7 took no Computer Science classes; 17 took 1-3; 15 took 4-6; 13 took 7-12, and 8 took over 13 classes.
11 took at least one class in Artificial Intelligence.
Participants were compensated via emailed \$10 Amazon gift card upon study completion.
We did not reject anyone who applied, as our criteria only included being an adult, color differentiation, and ability to use a computer.

\noindentbf{Study Conditions.}
We used a 2$\times$2 partial within-subject, partial between-subject design, to study the effects of the two variables
(i.e., nearest neighbors (NN) vs. cluster-to-cluster (CL) workflows; ImageNet (IM) vs. Density Ratio (DR)).
We randomly assign participants to the condition of using the ImageNet features (IM) or those learned by our Density Ratio (DR) covariate shift model (between-subjects). Both conditions used the same underlying CNN, density ratio model, and outlier scores.
Then, each participant performed two shift identification tasks, one with the NN workflow and the other with the CL workflow
(within-subjects).  
For example, a participant performed the first shift identification task 
(e.g., glasses, smile, and necktie) with the NN workflow that uses the ImageNet features (i.e., NN-IM), and then performed the second task 
(e.g., hats and facial hair) with the CL workflow that uses the same ImageNet features (i.e., CL-IM).
The condition orders were counter-balanced.

\noindentbf{Dataset.}
We used a subset of images from the CelebA faces dataset~\cite{liu2015faceattributes},
as non-experts can understand changing attributes on a face and would not need guidance on this part of the task.
We selected 6 attributes from the 40 potential attributes: eyeglasses, smiling, wearing necktie, wearing a hat, having a beard, and having a mustache. We separated these attributes into two sets of shifts: 
(1) glasses, smiles, and neckties; 
(2) hats and facial hair (beards/mustache combined). 
We selected 5,000 images as a training set, 9,000 as an unshifted test set, and 1,000 as a test set containing a shift.

\noindentbf{Study Procedure.}
Our study was conducted completely online, where participants performed on their own once provided website URL and login information.
The study begins with a video tutorial 
with an example 
that uses a toy dataset of flowers. 
Participants were asked attention check questions 
to ensure their understanding. 
Participants are then directed to perform the tasks with the interface for a minimum of 10 minutes and a maximum of 20 minutes for each shift set before submitting a Google form answer sheet of what they believe to be the sources of covariate shift (up to 5 responses).
The interfaces used in the study are shown in the supplemental material. 

\subsection{Study Data Collection and Analysis}

We coded participant responses to compare the number of participants who find each specific shift (e.g., eyeglasses) between conditions. For all statistical analysis we use a one-tailed 2$\times$2 Fisher's exact test. The contingency table of the Fisher's test is comprised of ``found specific shift'' vs. ``did not find it'' and condition ``a'' vs. ``b''. 
For example, we perform a test on the CL workflow with DR features vs. the same workflow with the IM features on the eyeglasses attribute.

\section{Results}

\begin{table}[!b] 
\centering

\begin{tabular}{cccccc}
\toprule
      Condition & Glasses & Smile & Necktie & Hats & Facial Hair \\ \midrule
NN-IM & 4       & 6     & 3                                                   & 14   & 6                                                      \\
CL-IM & 4       & 7     & 1                                                   & 14   & 1                                                      \\
NN-DR & 7       & \textbf{13}    & \textbf{4}                                                   & \textbf{15}   & \textbf{6}                                                      \\
CL-DR & \textbf{11}      & 7     & 1                                                   & \textbf{15}   & 2                                                      \\ 
\bottomrule
\end{tabular}
\vspace{-8pt}
\caption{The number of participants discovering each shift for all conditions, higher is better. The best performing method for each condition are highlighted in bold.}
\label{table:study_counts}
\end{table}

\begin{table}[!tb] 
\centering
\begin{tabular}{ r @{} l c @{\hskip7pt} c @{\hskip7pt} c @{\hskip7pt} c @{\hskip4pt} c @{\hskip2pt} }
\toprule
     \multicolumn{2}{c}{Comparison} & Glasses & Smile & Necktie & Hats & Facial Hair \\ 
\midrule
DR &\textgreater IM       & \textbf{0.01}    & 0.06  & 0.50                                                & 0.25 & 0.50 \\

NN-DR & \textgreater NN-IM & 0.22    & \textbf{0.01}  & 0.50                                                & 0.50 & 0.64                                                   \\
CL-DR &\textgreater CL-IM & \textbf{0.01}    & 0.64  & 0.76                                                & 0.50 & 0.50                                                   \\
\midrule
NN &\textgreater CL       & 0.22    & 0.15  & 0.07                                                & 0.75 & \textbf{0.01 }                                                  \\
NN-IM & \textgreater CL-IM & 0.66    & 0.77  & 0.30                                                & 0.76 & \textbf{0.04}                                                   \\
NN-DR &\textgreater CL-DR & 0.97    & \textbf{0.03}  & 0.16                                                & 1.00 & 0.11                                                   \\

\bottomrule
\end{tabular}
\vspace{-8pt}
\caption{A table of p-values for a comparison between a pair of conditions from the user study. Bolded numbers are statistically significant at $p < 0.05$ (one-tailed Fisher's Exact Test).
}
\label{table:study_statistic_tests}
\end{table}

\begin{table}[!tb]
\centering
\begin{tabular}{llll|lll @{}}
\toprule
      & \multicolumn{3}{c|}{Shift Set 1 } & \multicolumn{3}{c}{Shift Set 2 } \\
      & P $(\uparrow)$     & R $(\uparrow)$     & FP $(\downarrow)$       & P $(\uparrow)$      & R $(\uparrow)$     & FP $(\downarrow)$      \\ \midrule
NN-IM & 0.42           & 0.29        & 1.40      & 0.64            & 0.67        & 1.00      \\
CL-IM & 0.35           & 0.27        & 1.80      & 0.49            & 0.50        & 1.40      \\
NN-DR & {\bf 0.85}           & {\bf 0.53}        & {\bf 0.53}      & {\bf 0.70}            & {\bf 0.70}        & {\bf 0.93}      \\
CL-DR & 0.48           & 0.42        & 1.47      & 0.57            & 0.57        & 1.40      \\ \bottomrule
\end{tabular}
\vspace{-8pt}
\caption{Average precision (P), recall (R) and false positive rate (FP) by condition. The arrows indicate whether higher or lower is better.}
\label{table:study_statistics} %
\end{table}

Our results indicate that
the \textit{nearest neighbor} workflow with \textit{density ratio} latent representation (NN-DR) is generally the best combination for identifying covariate shift. Table \ref{table:study_counts} shows the count of the number of participants who find a specific shift. Table \ref{table:study_statistic_tests} shows the p-values from Fisher's test statistic for the comparison of each condition to another. We find NN-DR is significantly better than CL-DR as well as CL-IM at finding smiles. Other than for detecting eyeglasses, no method performs better than NN-DR; even on eyeglasses, the improvement by the cluster view is not statistically significant.

When comparing the conditions by representation, the density ratio representation is always equivalent to or better at identifying covariate shift than the original ImageNet space. 
Some shifts lent themselves to a specific workflow rather than a latent space. Finding neckties and facial hair is difficult for cluster workflow users, with only two and three participants finding it, respectively. However, using the nearest neighbor workflow is nearly statistically significant for neckties and is significant for finding facial hair (one-tailed Fisher's Exact Test, $p<0.05$). We believe the primary benefit to users is the ability to focus on a single image to compare and contrast both the train and the test sets. In the one instance where NN-DR is not the best (for detecting eyeglasses), the CL-DR is not statistically significantly superior (one-tailed Fisher's Exact Test, $p=0.132$).

The cluster workflow is also more detrimental to participants finding shifts that do not exist. Table \ref{table:study_statistics} shows the average precision, recall, and false positive rate for each condition.  NN-DR is the highest in precision and recall and the lowest in false positive rate. In terms of user workflows, the precision for the nearest neighbor workflow is at least 0.37 higher for shift set 1 and 0.13 higher for shift set 2 when compared to the cluster workflow. These results suggest that by having a focal image, participants are better able to compare and contrast the training and test sets.

\begin{figure}[!tb]
\centering
        \includegraphics[width=.92\linewidth]{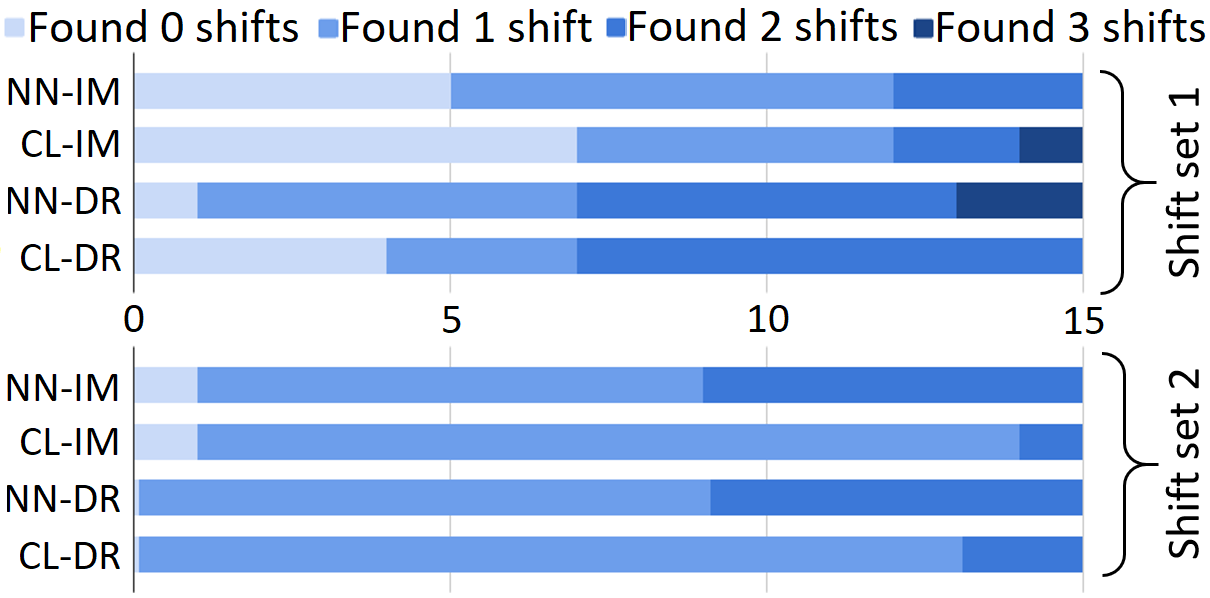}
        \vspace{-8pt}
        \caption{
        Number of participants on how many shifts they found among three (for shift set \#1) or two different shifts (for shift set \#2). For example, for the NN-DR condition for shift set \#1, 
        14 out of 15 participants found at least one shift and two found all the three shifts.} 
        \myvspace
        \label{fig:multi_detect_histo1}
\end{figure}

Lastly, participants in the nearest neighbor workflow are consistently better at finding more shifts (e.g., finding one, two, and all three shifts from shift set 1). Breakdowns  are shown in \autoref{fig:multi_detect_histo1}.

\section{Discussion}
Our results clearly point toward two outcomes: the importance of selecting an appropriate latent representation and a user's need to have a focal image for which to compare against a group. Our cluster workflow generally performed worse than we originally expected, and as Participant \#20 said, ``I believe having a selected image helps to understand the process [of finding the shift] better''. 
It is clear that not having a central selected image makes the shift detection task much harder, as a user must compare all test images in a given cluster to each other and also to all the images in the training set. %

The nearest neighbor workflow was not always able to outperform the cluster workflow, such as for detecting eyeglasses. We speculate  if participants had more time, and thus more examples of outliers than the limited few in the top 100 images, 
they may have been better able to identify that shift. The density ratio space cluster view (CL-DR) created a side-by-side histogram of all face images wearing sunglasses in the test set, which no other condition generated.

A final result to note is that at least one user from all conditions found each shift. This finding validates our experimental setup was not biased in favor of or against a particular condition.

\section{Conclusion and Future Work}

This work is one of the first to investigate analytic workflows for detecting covariate shifts in image data and to investigate the effect of latent representations on how well human users detect them. Our results indicate that using a nearest neighbor approach combined with a density ratio latent representation enabled participants to accurately discover and characterize different types of localized covariate shift. 

While our results are promising,
we want to note the limitations of this work.
The main caveat is naturally the limitations of the data itself. 
We used a relatively small dataset that exhibits covariate shift. We leave it for future work to examine cases in extremely large data settings, or in settings where no covariate shift has occurred.

\acknowledgments{
This work was supported by DARPA \#N66001-17-2-4030. 
}

\clearpage
\bibliographystyle{abbrv-doi}
\bibliography{references}
\clearpage

\textbf{\textsc{\LARGE Supplementary Material}}
\appendix
\maketitle
\section{Performance of Machine Learning Techniques to Detect Covariate Shift Instances}
This section provides more details regarding the experiments in Section 3 involving the latent representation and  scoring function combinations. 

In order to help users detect test instances affected by covariate shift, we need to decide on which latent representation to use, as well as how to score the test instances as outliers. Many different latent representations are possible for our user study. To inform our decision, we run an experiment to determine which combination is most effective for a machine learning method to detect outliers. For our experiment, we evaluate a large combination of latent space representations with different scoring functions. 

We briefly describe some of these combinations.  For latent representations, we investigate using the latent space of a classifier trained on a separate dataset (i.e., the ImageNet dataset of over 1 million images with 1,000 classes). This transfer learning approach is an effective technique that is commonly used in many image-related tasks \cite{huh2016makes}. In addition, we explore using latent spaces generated by the VGG11 classifier \cite{simonyan2014very}, autoencoders (AE) \cite{Hinton94} and variational autoencoders (VAE) \cite{Kingma14}. The fourth latent representation we explore is learned by direct density ratio estimation \cite{Nam15}, specifically using the Kullbeck-Liebler Importance Estimation Procedure (KLIEP). KLIEP returns the importance estimate $r(X) = P^{\text{te}}(\bm{X}) / P^{\text{tr}}(\bm{X})$, which is the ratio of the test density divided by the training density. The higher the ratio, the more likely the instance will be an outlier.
We also experimented with passing in other latent representations (e.g., ImageNet) to KLIEP instead of the original input features; this is indicated as +KLIEP in Table \ref{tab:latent_representations_complete}.

For scoring functions, we investigated the use of reconstruction loss, 1 - $P(Y|\bm{X})$, and we also applied an effective anomaly detection algorithm called Isolation Forest \cite{liu2008isolation} to the latent representation. In the case of density ratio estimation, we score potential outliers by their importance estimate.

For this experiment, we use an image dataset (CelebA \cite{liu2015faceattributes}) consisting of over 200 thousand celebrity faces with 40 labeled attributes each. 
We take a subset of the training data where a given attribute is present (or absent), but where the test set still contains the original attribute. This gives us 80 different covariate shift experiments where a method can be tested to see how well it identifies the shifted images in the test set. 

Table \ref{tab:latent_representations_complete} summarizes the AUROCs for the different approaches over the 80 experiments. Due to space limitations, we only include a subset of the results but the full set of results can be found in Section A of the supplementary material. Passing other latent representations to KLIEP produces large gains in performance, indicating the benefit of using density ratio loss functions. The best performing combination was the latent representation learned when a pre-trained ImageNet representation was passed as input to the KLIEP model and a scoring function based on density ratio; we refer to this combination as the \emph{density ratio latent representation} in latter sections.

With these results to inform our user study, we use a pre-trained ImageNet CNN as a baseline and compare it against a density ratio latent representation.

Unless noted otherwise, all trained models iterate for 30 epochs over the training set using an Adam \cite{kingma2014adam} optimizer with default parameters and a learning rate of $0.001$. We next describe the different latent space representations that we use in our experiments below.

\begin{table}[!t]
\centering
\begin{tabular}{llc}
\toprule
\bf{Latent representation} & \bf{Scoring function}                     & \bf{AUROC} \\ \midrule
VGG11 Classifier      & 1-P(Y|X)                               & 0.55  \\
VAE                   & Latent dist. from ctr                & 0.53  \\
AE                    & Reconstruction Error                 & 0.50  \\
VAE                   & Reconstruction Error                 & 0.54  \\
VGG11 Classifier      & Isolation Forest                     & 0.54  \\
AE                    & Isolation Forest                     & 0.51  \\
VAE                   & Isolation Forest                     & 0.51  \\
KLIEP                 & Isolation Forest                     & 0.52  \\
ImageNet              & Isolation Forest                     & 0.54  \\
VGG11 Classifier+KLIEP & Density Ratio                        & 0.69  \\
AE+KLIEP              & Density Ratio                        & 0.73  \\
VAE+KLIEP             & Density Ratio                        & 0.74  \\
KLIEP                   & Density Ratio                        & 0.57  \\
{\bf ImageNet+KLIEP}        & {\bf Density Ratio}                        & {\bf 0.80}  \\ \bottomrule
\end{tabular}
\vspace{-5pt}
\caption{The complete set of results for latent space representations and scoring function combinations in our experiments}
\vspace{-2pt}
\label{tab:latent_representations_complete}
\end{table}

\begin{itemize}[itemsep=0pt]
\item {\bf VGG11 Classifier.}
Using the probability of a predictive class is a common technique for finding uncertain test examples \cite{hendrycks2016baseline}. We train a classifier $M(\bm{x})$ with a VGG11 architecture to identify an attribute of interest $y$ for a given image $\bm{x}$. For our experiments we use the ``Male'' attribute as it splits the dataset the most evenly. We define the latent representation of this classifier to be $\bm{z} = M(\bm{x})$ and the predicted classification to be $\hat{y} = \bm{W}^T \bm{z} + b$ where $W$ and $b$ are learned parameters with size $(d \times 1)$ and $1$ respectively.

\item {\bf Pretrained ImageNet Classifier.}
We take a classifier trained on the ImageNet dataset (of over 1 million images with 1000 classes) and use its latent representation $z = M(x)$. This transfer learning approach is an effective dimensionality reduction technique that is commonly used in many image-related tasks \cite{huh2016makes}. Specifically, we use the pretrained InceptionNet \cite{szegedy2015going} architecture.

\item{\bf Auto-Encoder.}
Our auto-encoder \cite{Hinton94} (AE) is a deep convolutional neural network comprised of two parts: an encoder and a decoder. The encoder $E(x)$ takes an image $x$ as input and reduces the dimensionality to a relatively small real-valued vector $z = E(x)$. 
The decoder $D(z)$ is a deconvolutional neural network which takes the compressed representation $z$ and outputs an image with the same dimensions as $x$ which we call $\hat{x} = D(z)$. 
An auto-encoder is optimized to minimize the difference between $x$ and $\hat{x}$ for all images in the training dataset, $L_{\text{AE}} = \sum_{x \in \bm{X}} ||x - D(E(x))||^2$.
Reconstruction error is a commonly used metric for determining if images belong to the train set as non-members should auto-encoder poorly \cite{zhou2017anomaly}.

\item{\bf Variational Auto-Encoder.}
A Variational Auto-Encoder \cite{Kingma14}  (VAE) has the same setup as a regular auto-encoder except with an additional loss function term of $D_{\text{KL}}(E(x) || p(z))$ where  $p(z)$ is a multi-variate Gaussian prior with the same dimensionality as $z$ , $p(z) = \mathcal{N}(0,1)$.
Therefore the loss of the VAE is $L_{\text{VAE}} = L_{\text{AE}} + \sum_{x \in \bm{X}} D_{\text{KL}}(E(x) || p(z))$.

\item{\bf KLIEP latent space.} 
We also used the latent space learned by direct density ratio estimation \cite{Nam15}, specifically using the CNN version of the Kullbeck-Liebler Importance Estimation Procedure (KLIEP). In cases where we provide a latent representation (e.g. ImageNet) instead of the original features as an input to KLIEP, we use a 2-layer multi-layer perceptron trained with the KLIEP loss for 10 epochs using stochastic gradient descent and a learning rate of $0.01$. We use the +KLIEP extension to indicate this combination.

\end{itemize}

For scoring functions, we use the following:
\begin{itemize}[itemsep=0pt]
\item {\bf 1-Class probability.} We use the probability $1-P(Y|\bm{X})$ as the scoring function for an outlier
\item {\bf Distance from latent space center.} For VAEs, we can use the distance of the test instance from the center of the latent space as an outlier score.
\item {\bf Reconstruction error.} For AEs and VAEs, we can use the reconstruction error as the outlier score.
\item {\bf Isolation Forest.} The Isolation Forest algorithm \cite{liu2008isolation} is an unsupervised ensemble-based anomaly detection technique which identifies outliers as points that are easily isolated by random splits. We use the builtin sklearn implementation for this algorithm. We generate a forest for all inputs from both training and test set, then use the isolation path length as an anomaly score. Given the latent representations of test instances, we can apply Isolation Forest to detect outliers.
\item {\bf Density ratio.} We score potential outliers by their density ratio $\frac{P^{\text{te}}(\bm{X})}{P^{\text{tr}}(\bm{X})}$, with the higher the ratio, the more likely an outlier the instance is.  
\end{itemize}

Table \ref{tab:latent_representations_complete} contains a summary of the results.

\begin{figure*}[!tb]
\centering
        \includegraphics[width=.8\paperwidth]{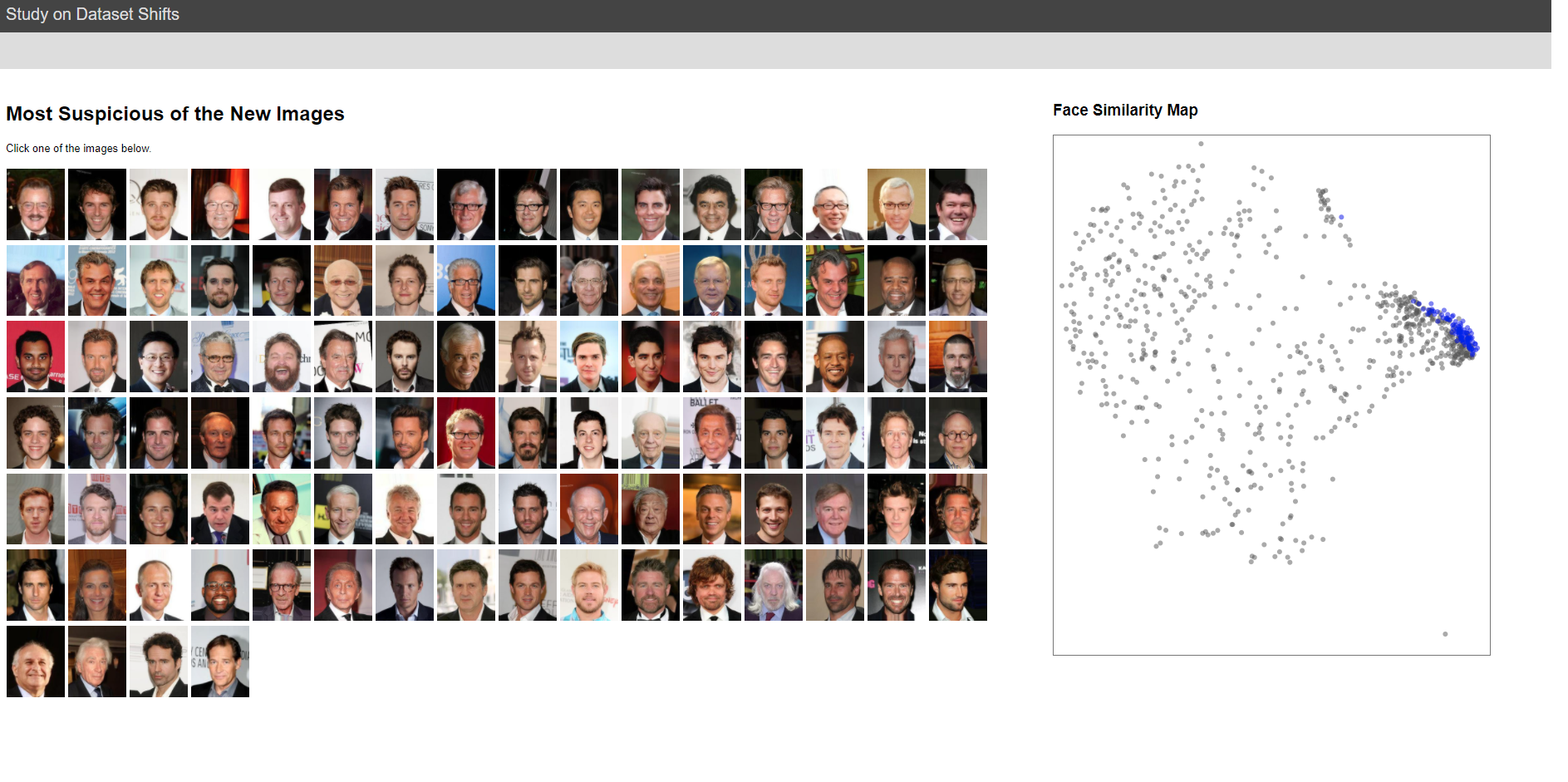}
        \caption{An example of the front page for the nearest-neighbors workflow.
        }
        \label{fig:full_nn_home}
\end{figure*}

\begin{figure*}[!tb]
\centering
        \includegraphics[width=.8\paperwidth]{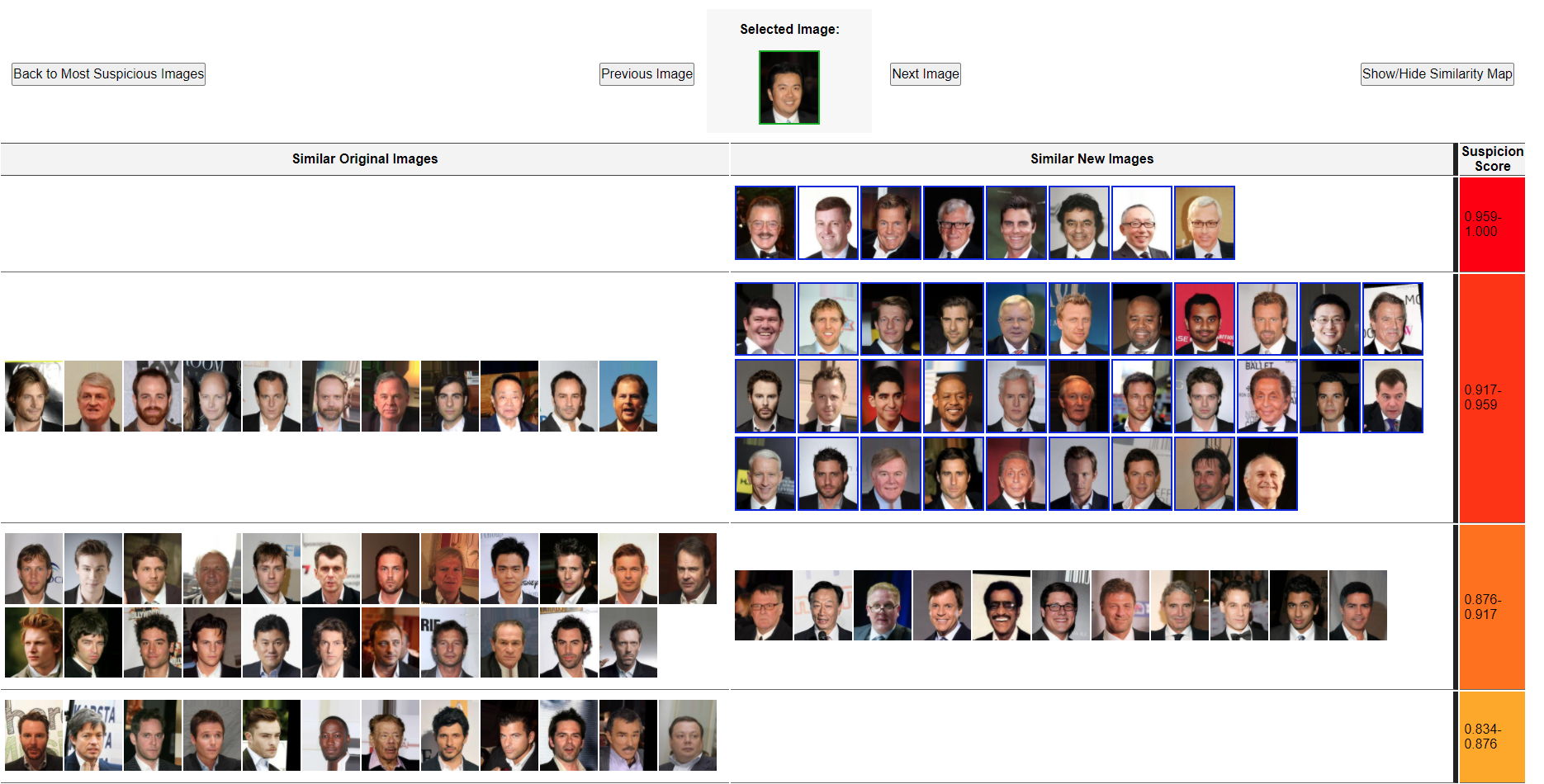}
        \caption{An example of the \textit{side-by-side histogram} from the nearest-neighbors workflow.
        }
        \label{fig:full_nn_histogram}
\end{figure*}

\begin{figure*}[!tb]
\centering
        \includegraphics[width=.8\paperwidth]{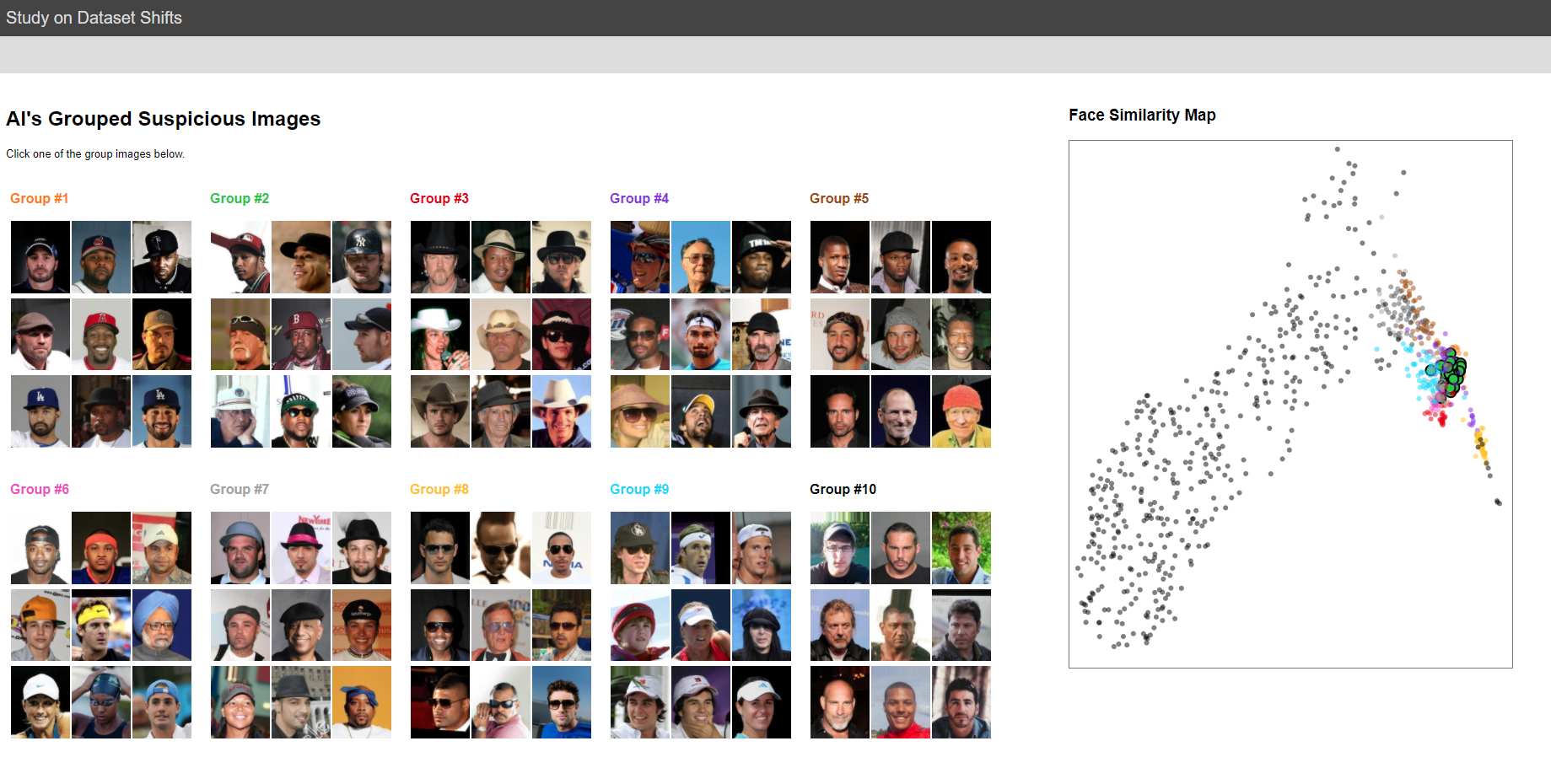}
        \caption{An example of the front page for the cluster-to-cluster workflow.
        }
        \label{fig:full_cluster_home}
\end{figure*}

\begin{figure*}[!tb]
\centering
        \includegraphics[width=.8\paperwidth]{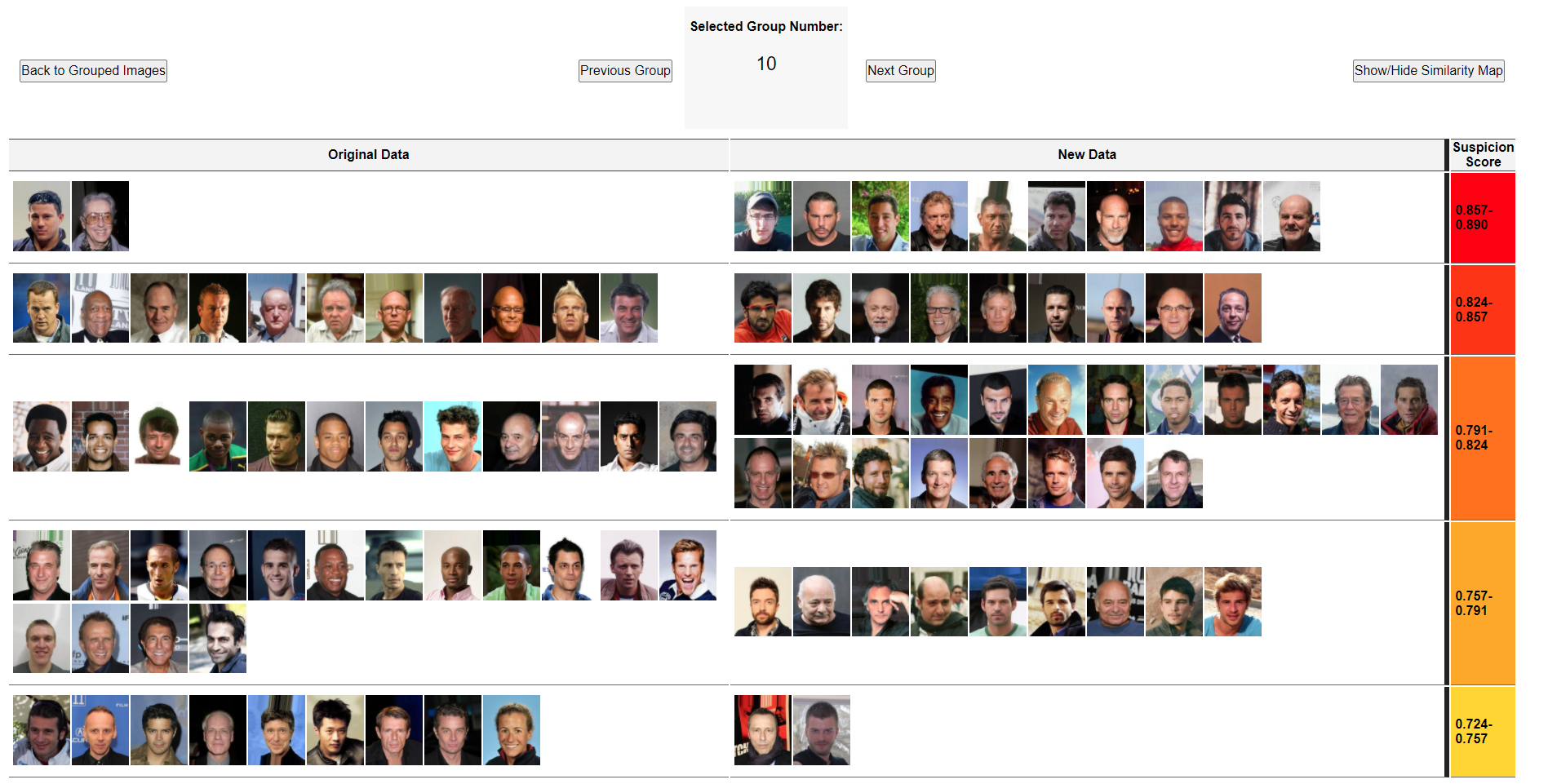}
        \caption{An example of the \textit{side-by-side histogram} from the cluster-to-cluster workflow.
        }
        \label{fig:full_cluster_histogram}
\end{figure*}

\section{User Interfaces for User Study}

This section presents the full user interfaces used in our user study. Figures \ref{fig:full_nn_home} and \ref{fig:full_cluster_home} show the front pages for the nearest-neighbor and cluster-to-cluster workflows, respectively. 
Figures \ref{fig:full_nn_histogram} and \ref{fig:full_cluster_histogram} show the \textit{side-by-side histogram} for the nearest-neighbor and cluster-to-cluster workflows, respectively.

\end{document}